# Scalable Relaxations of Sparse Packing Constraints: Optimal Biocontrol in Predator-Prey Networks

**Johan Bjorck**
Dept. of Computer Science
Cornell University
njb225@cornell.edu

**Yiwei Bai**
Dept. of Computer Science
Shanghai Jiao Tong University *
yb263@cornell.edu

**Xiaojian Wu**
Dept. of Computer Science
Cornell University
xw458@cornell.edu

**Yexiang Xue**
Dept. of Computer Science
Cornell University
yx247@cornell.edu

**Mark Whitmore**
Dept. of Natural Resources
Cornell University
mark.whitmore@cornell.edu

**Carla Gomes**
Dept. of Computer Science
Cornell University
gomes@cs.cornell.edu

## Abstract

Cascades represent rapid changes in networks. A cascading phenomenon of ecological and economic impact is the spread of invasive species in geographic landscapes. The most promising management strategy is often biocontrol, which entails introducing a natural predator able to control the invading population, a setting that can be treated as two interacting cascades of predator and prey populations. We formulate and study a nonlinear problem of optimal biocontrol: optimally seeding the predator cascade over time to minimize the harmful prey population. Recurring budgets, which typically face conservation organizations, naturally leads to sparse constraints which make the problem amenable to approximation algorithms. Available methods based on continuous relaxations scale poorly, to remedy this we develop a novel and scalable randomized algorithm based on a width relaxation, applicable to a broad class of combinatorial optimization problems. We evaluate our contributions in the context of biocontrol for the insect pest Hemlock Wolly Adelgid (HWA) in eastern North America. Our algorithm outperforms competing methods in terms of scalability and solution quality and finds near-optimal strategies for the control of the HWA for fine-grained networks – an important problem in computational sustainability.

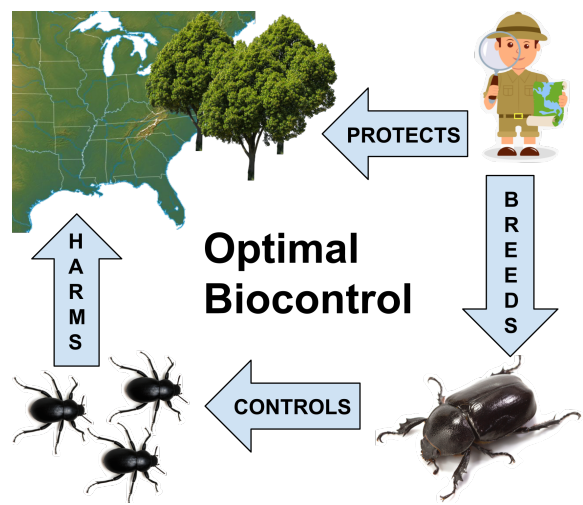

Figure 1: The insect pest HWA has recently invaded hemlock forests of eastern North America. By breeding and introducing a natural predator conservationists can control the invasive population (Onken and Richard 2011). The challenge of optimal biocontrol is using available biocontrol resources across large and complicated ecosystems for maximum impact.

## Introduction

Cascades in networks model settings such as idea propagation in social networks (Kempe, Kleinberg, and Tardos 2003) or the spread of infectious diseases (Aspnes, Chang, and Yampolskiy 2005). Of economic and ecologic importance is the cascading behavior of invasive species invading local ecosystems, two famous examples of invasives species are the cane toad in Australia or the Nile Perch in Lake Victoria, both having quickly devastated local ecology (Goldschmidt, Witte, and Wanink 1993) (Burnett 1997). Invasive species cause billions of dollars in economic damage (Pimentel, Zuniga, and Morrison 2005) and the most promising management strategy is often **Biocontrol** (Onken and Richard 2011), which encompasses releasing a natural predator that is able to control the invasive population. Conservation organizations face the challenge of judiciously using available biocontrol resources to minimize the ecological damage of harmful invading species, a problem we aim to study from a computational perspective.

Drawing upon models from ecological literature we **formulate a novel problem** of optimal biocontrol. Predator and prey populations are modeled as interacting cascades on a graph, and the biocontrol problem entails optimally seeding the predator cascade over time to minimize the invading prey population. Conservation organizations typically operate over long periods of time subject to yearly resource budgets; this is modeled as a number of resources (e.g. money, available predators, equipment) with individual yearly budgets. We show that the resulting biocontrol problem is NP-

---

hard, suffers from diminishing returns and how recurring budgets naturally introduce constraint sparsity. These characteristics are prevalent in real-world combinatorial problems (Kallrath and Schreieck 1995) (Leskovec et al. 2007), and the problem can be stated as one of submodular optimization subject to sparse knapsack constraints.

In general, greedy strategies provides no approximation factor to this broad problem class (see Figure 3). A continuous relaxation pioneered in (Calinescu et al. 2007) can provide an approximation guarantee that is linear in a 'sparsity parameter' (Bansal et al. 2012), which is much better than competing methods for sparse problems. The method suffers from poor scalability and to remedy this we develop a new, **faster approximation algorithm for sparse submodular optimization**. Our algorithm is based on a width relaxation and a randomized projection step, which enables fast primal-dual techniques to be used. This circumvents the $\mathcal{O}(n^5)$ sampling step of the continuous relaxation and the algorithm instead runs in $\mathcal{O}(n^2)$.

As part of an ongoing collaboration with the Department of Natural Resources at Cornell we evaluate our techniques in the context of biocontrol of the invasive insect pest Hemlock Wolly Adelgid (HWA), see Figure 4. Using HWA models from ecological literature (Fitzpatrick et al. 2012) (Hakeem et al. 2013) and geographical datasets we study a **realistic instance of citizen science-based biocontrol** in eastern North America. Our algorithm is able to solve the problem in a fine-grained network close to optimality, which is needed for precision biocontrol. Our contributions lie in 1) **introducing a novel problem of biocontrol**, applicable for unevenly competitive cascades broadly, and showing fundamental properties of the problem. 2) **Developing a new approximation algorithm for sparse submodular optimization** able to scale vastly better than competing methods by the use of a novel relaxation scheme. 3) **Evaluating our algorithm on HWA biocontrol, an important problem in computational sustainability** (Gomes 2009). Here our algorithm vastly outperforms both other approximation algorithms and heuristic methods used by ecologists, and points towards non-trivial management strategies.

## Problem Formulation

### A General Predator-Prey Model

Our general predator-prey framework is based on metapopulation models (Hanski 1999), which describe the population dynamics of habitat patches in a landscape, see Figure 2. The geographic landscape the predator and prey spread in is discretized into a number of patches of land, each one represented by a vertex in a weighted directed graph $G = (V, E)$. Edges correspond paths along which the species can spread. The spreading process unfolds at discrete time steps under the following assumptions

- Every node can be in one of **three states** - unoccupied, prey or predator. The predator is assumed to feed only on the prey, hence the presence of the prey is a prerequisite for establishing and sustaining a predator population. The predator state implies a balanced population, the predator keeping the prey population below damaging levels.

- The prey and predator both spread according to a **cascade model**. Unoccupied nodes connected to a predator or prey node can become occupied by the prey as a product of diffusion or migration from the predator. Nodes are required to be in the prey state before entering the predator state.

These assumptions are similar to the ones of classical continuous predator-prey metapopulation models (Hastings 1977), like them we assume discrete states and the state change order unoccupied → prey → predator. The initial conditions are given by disjoint sets $O$, $P$ and $Q$ that represent nodes in the unoccupied, prey and predator states respectively at time $t = 0$. At any subsequent time step $t$ an edge $(u,v)$ between an unoccupied vertex $u$ and any vertex $v$ not unoccupied allows the prey to spread from $v$ to $u$ with probability $p_{(u,v)}$. If at least one neighbor $v$ spreads the prey to $u$ at time step $t$, the node $u$ is in the prey state at time $t + 1$, otherwise, it remains in the unoccupied state. Once in the prey state a node $u$ stays that way until it enters the predator state. The predator spreads identically the prey, but from nodes in predator state to nodes in the prey state, with probability $p'_{(u,v)}$. Models of biological processes quickly can become computationally intractable, for example, Conway's game of life is Turing complete (Rendell 2002). Our relatively simplistic assumptions avoid this, and while they aren't universally applicable, in our application of biocontrol of the HWA across the eastern North America the predator is known to only feed on the prey (Cheah and McClure ) and be able to coexist with it across large habitats (Sasaji and McClure 1997).

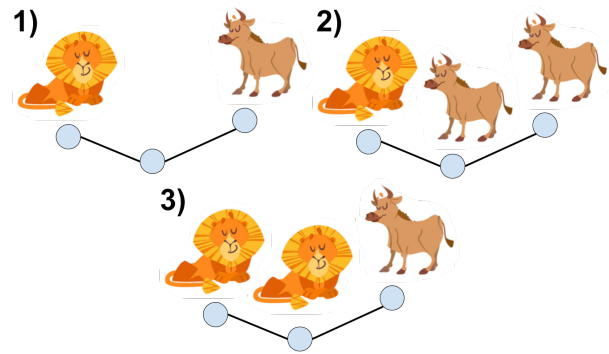

Figure 2: The predator and prey populations spread according to a cascade model. The central node is initially unoccupied, this enables the prey (Buffalo) to spread to it. Only after the prey has spread to the central node the predator (Lion) is able to spread further by feeding on the new prey population.

### Problem Definition

We now formulate the problem of optimal biocontrol - introducing the predator to minimize the damage caused by a harmful prey population. Conservation organizations typically operate over long periods of time subject to yearly resource constraints, for example, yearly operating budget or man hours. We assume that there are $k$ **types of resources** (e.g. equipment, available predators, personnel, money), and that there are individual budgets at every time step for those resources. The cascading process unfolds in a finite number of steps $T = \{0, 1, ..., T_{max}\}$ and at $t = 0$ we wish to choose a strategy $X \subseteq V \times T$, specifying where and when to release the predator in future time steps, subject to these constraints. The predator is introduced according to the locations and times of the strategy, and if food (prey) is available the land patches enter the predator state. The resulting cascades are random processes, we let $P_t(X)$ denote the random set of nodes in $G$ that are in the prey state at time $t$, when predators are released according to strategy $X$. We now introduce $f(X)$, the objective function we will aim to maximize subject to our constraints.

$$f(X) = \mathbb{E}\left[\sum_{t \in T} |P_t(\emptyset)| - |P_t(X)|\right] \quad (1)$$

The function $f(X)$ corresponds to the number of nodes we save from infestation by the prey through strategy $X$, and how early we save them, in expectation. Any item in the set $V \times T$ has fixed resource costs, and every time step in T has individual budgets for each of the k resources that cannot be exceeded. We hence have $|T| \times k$ constraints in total and can describe the resulting linear constraint by a nonnegative matrix $A$ and vector $b$. These two represent the costs for different elements in $V \times T$ and the yearly budgets respectively. Any action taken at time $t$ just consumes the resources allotted for year $t$ - it's not possible to use next year's resources for this year's management actions. Hence for any column $j$ in $A$, which represents the costs for action $j$ in various constraints, at most $k$ entries, corresponding to budgets for various resources at year $t$, are non-zero. We say that $A$ is **k-column sparse**. We will interchangeably use vector and set notation for items in the packing problem, statements like $AX \leq b$ or $c^T X$ is to be interpreted by exchanging the set $X$ by its corresponding vector $\in \{0, 1\}^{|V \times T|}$. We can now write our budget-constrained optimization problem as

$$\max_{X \subseteq V \times T} f(X) \quad s.t. \quad AX \leq b \quad (2)$$

## Algorithms

Problem (2) is a combinatorial optimization problem, and its NP-hardness as per proposition 1 is perhaps not surprising. A formal proof is given in online version (Bjorck et al. 2017), at a high level it constructs graphs such that the problem expresses the maximum set cover problem (Garey and Johnson 2002).

**Proposition 1** *Problem* (2) *is NP-hard.*

As common in many real world problems, see for example (Leskovec et al. 2007) (Badanidiyuru et al. 2014), problem (2) exhibits diminishing returns. The marginal utility of introducing predators decreases with the amount of already introduced predators. From an ecological perspective this is natural; predator colonies eventually start competing for resources. A function $f$ is *submodular*, if for any item $i$, and sets $A$ and $B \subseteq A$ with $i \notin A$, we have $f(A \cup \{i\}) - f(A) \leq f(B \cup \{i\}) - f(B)$.

**Proposition 2** *The objective function $f(X)$ is monotonically submodular.*

Submodularity of the objective function $f$ stems from our assumptions about the predator and prey population dynamics. We have assumed that the predator only feeds on the prey, and hence it can only spread to where the prey has previously spread. Thus, the predator cascade must "follow in the footstep" of the prey cascade, and the different predator populations risk spreading to reach the same parts of the prey cascade where their overlap adds no benefit. The formal proof is given in the online version of this paper.

With proposition 2 and our assumptions about $k$ resources with yearly budgets, problem (2) reduces to **maximizing a monotonic submodular function subject to $k$-column-sparse packing constraints**. We'll denote this problem by SUBSPPACK. The term "packing constraints" simply refer to nonnegative linear constraint. This general problem class subsumes prototypical combinatorial problems (e.g. knapsack and sparse independent set) and various submodular optimization problems that arise in applications (Leskovec et al. 2007) (Badanidiyuru et al. 2014) (Hoi et al. 2006). We will hereafter consider this general class of problems, and following standards in submodular optimization (Badanidiyuru and Vondrák 2014) we assume that $f(X)$ accessed through an oracle and measure algorithm complexity in the number of oracle calls (evaluating $f$ often becomes the bottleneck in practice). Additionally, we let $U$ denote all items (numbered $u_1, u_2..u_n$) that can be selected, $m$ the total amounts of constraints and $n$ the size of the set $f(X)$ is defined on. $A$ is then $m$-by-$n$, and we scale $A$ and $b$ WLOG such that $b = 1$. We assume that any single item constitutes a feasible solution, $m > 1$ (for $m = 1$ use (Sviridenko 2004)) and that all items do not fit the relaxed knapsack (if so take them all).

For this class of optimization problems greedy strategies provide no guarantees, see Figure 3 for a case with unbounded approximation ratio. In fact, SUBSPPACK cannot be approximated much better than to a factor $\mathcal{O}(k)$ as it contains the k-set packing problem as a special case, which cannot be approximated within a factor $\Omega(k/\log(k))$ assuming $P \neq NP$ (Hazan, Safra, and Schwartz 2006). Selecting a management strategy for our biocontrol problem dynamically, i.e. allowing actions to be taken in response to how the cascades spread, also contains the k-set packing problem – hence a dynamic policy essentially cannot improve the problem approximability. An approximation factor $\mathcal{O}(k)$ is actually achievable, which for sparse problems is much better than the $\mathcal{O}(m)$ approximation that generic methods achieve, see for example (Badanidiyuru and Vondrák 2014). The only algorithm known to the authors that achieves a $\mathcal{O}(k)$ guarantee relies on sampling a **continuous relaxation** (Bansal et al. 2012), but as this sampling uses $\mathcal{O}(n^5)$ function evaluations this algorithm quickly becomes impractical.

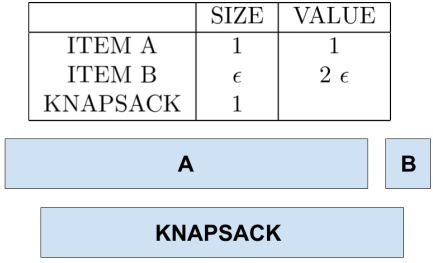

Figure 3: In the knapsack problem illustrated above, a greedy algorithm sorting items by value/size will fail – it would first select item $B$ and would then not be able to fit the more valuable item $A$. Our solution is to make the knapsack twice as large, pack it with the greedy value/size strategy, and finally randomly remove half of the items. With minor fixes, this constitutes an approximation algorithm that scales much better than continuous relaxations for submodular functions. Algorithm 1 and 2 extend this idea to higher dimensions.

## A Novel Relaxation Scheme

To solve SUBSPPACK in a scalable manner, we develop a novel relaxation-based algorithm. Fast algorithms for submodular optimization often rely on greedy strategies, however with general packing constraints (rather than cardinality constraints) greedy strategies fails. Figure 3 illustrates how a greedy choice makes it impossible to later fit a large and valuable item. By relaxing (2) through increasing our budget by some constant factor, a single greedy choice will not make it impossible to chose a large item later. This strategy avoids the pitfalls depicted in Figure 3, and enable scalable greedy algorithms.

**Definition 1** *The $\gamma$-width relaxation with $\gamma \geq 1$ is defined for packing constraints $Ax \leq b$ as the relaxed constraints $Ax \leq \gamma b$.*

Once we have solved the relaxed problem, we can obtain a solution to the original problem by randomly throwing items away. It is possible to illustrate the virtue of this strategy in case of the simplest packing problem: the knapsack problem. For any knapsack problem $K$, the optimal strategy to its fractional relaxation $K_{frac}$, where we can choose fractional items, is a greedy value/size-strategy (Cormen 2009). Assume that we pack a 2-width relaxation of $K$, let's call it $K_{wide}$, greedily. At the point when the size of the solution surpasses the original knapsack's size, the total value of the packed items must be at least as large as $\text{OPT}(K_{frac})$ since both strategies are identical up to that point. Hence $V_{wgreed}$, the value of the greedy solution to $K_{wide}$, satisfies $V_{wgreed} \geq \text{OPT}(K_{frac}) \geq \text{OPT}(K)$. By randomly removing items, the solution 'fits in expectations' and by linearity of expectation it achieves a constant factor approximation (again, in expectation).

These ideas are generalized to higher dimensions and made more precise in Theorem 1, 2 and 3. We show that the width relaxation of SUBSPPACK for $\gamma = \beta \log(m)$, for sufficiently large $\beta$, can be solved up to a constant factor with Algorithm 1. A feasible solution to the original problem is then constructed from the relaxed solution by randomly removing items in Algorithm 2, extending techniques from randomized LP rounding to the context of width relaxations of submodular programs (Bansal et al. 2012). The resulting method runs $\mathcal{O}(n^3)$ faster than the continuous alternative and gives a guarantee just a logarithmic factor away. Our method is presented in Algorithm 1 and 2, by combining theorem 2 and 3 we have the following theorem, see the proof at the end of the section.

**Theorem 1** *Algorithm 1 and 2 together form a randomized approximation algorithm for SUBSPPACK with an approximation factor of $\mathcal{O}(k \log(m))$ using $\mathcal{O}(n^2)$ function evaluations.*

## Algorithms for Sparse Packing

Algorithm 1 maintains a dual variable $y$, intuitively it can be thought of as the price for the 'resources' associated with every constraint. It increasing the price for resources in high demand using the $(1 + \epsilon)$ updates of the hedge algorithm (Kleinberg 2007) as in (Plotkin, Shmoys, and Tardos 1995). At each iteration, the marginal utility of adding any item to the solution is observed, and the algorithm greedily picks the item with the highest utility per "cost". The algorithm terminates when it chooses an item that cannot feasibly be added to the solution. This strategy forms an approximation algorithm to the $\gamma$-width relaxed SUBSPPACK problem for suitable $\gamma$, see Theorem 2.

---
**Algorithm 1** RelaxedDualPacking($\gamma, \epsilon$)
---
1: $S_0 \leftarrow \emptyset, y_i \leftarrow 1 \ \forall i = 1, 2...m$
2: **for** t = 0, 1,2... **do**
3: $\quad c_t^{(i)} = f(S_t \cup \{u_i\}) - f(S_t) \ \forall i = 1, 2...n$
4: $\quad r_t \leftarrow \mathbf{argmax}_{x \in U \setminus S_t} \frac{c_t^T x}{y^T Ax}$
5: $\quad$ **if** $A(S_t \cup \{r_t\}) > \gamma$ **then**
6: $\qquad$ **Return** $S_t$
7: $\quad$ **else**
8: $\qquad S_{t+1} \leftarrow S_t \cup \{r_t\}$
9: $\qquad y_i \leftarrow y_i(1 + \epsilon)^{A_{it}}, \quad \forall i \in \{1, 2...m\}$
10: $\qquad y \leftarrow y/\|y\|_1$
---

**Theorem 2** *Algorithm 1 is a constant factor approximation algorithm to the $\gamma$ width relaxatation of SUBSPPACK for suitable $\gamma = \mathcal{O}(\log(m))$ and $\epsilon$.*

**Proof.** Let OPT denote the optimal value of the $\gamma$-width relaxed SUBSPPACK problem, $S^*$ the optimal set, $S$ the set the algorithm produced and define $\gamma = \beta \log(m)$. If the algorithms output and the optimal solution overlaps such that $f(S \cap S^*) \geq \text{OPT}/3$, we clearly have a constant factor approximation. Assuming that's not the case we must have $f(S^* \setminus S) \geq 2 \text{ OPT}/3$ by submodularity. Additionally, if $f(S) \geq \text{OPT}/3$ we again have a constant factor approximation factor. If neither of those conditions hold we can use the trivial equation $f(A \cup B) - f(B) \geq f(A) - f(B)$, taking $A = S^* \setminus S$ and $B = S$. That gives us the equation $f(S \cup (S^* \setminus S)) - f(S) \geq \text{OPT}/3$, which in turn implies

(by submodularity) that $f(S_t \cup (S^* \setminus S)) - f(S_t) \geq \text{OPT}/3$ for every $t$. Now, by submodularity and the definition of $c_t$ according to line three in Algorithm (1), we have

$$c_t^T(S^* \setminus S) \geq \text{OPT}/3 \qquad \forall t \qquad (3)$$

This expression allows us to use regret-bounds similar to LP-packing, but as we cannot decrease our step size arbitrarily we will have to handle error terms differently. We define $V_{alg} = \sum_t c^T r_t$, the total value of our solution, and by $S'$ denote the vector $\in \{0,1\}^n$ corresponding to $S^* \setminus S$. As in (Plotkin, Shmoys, and Tardos 1995) we also define the weighted average dual variable

$$\bar{y} = \frac{1}{V_{alg}} \sum_t (c^T r_t) y_t$$

Since $S' \subseteq S^*$, $S'$ is feasible in every constraint, and hence

$$\gamma \geq \bar{y}^T A S' = \frac{1}{V_{alg}} \sum_t (c^T r_t)(y_t^T A S') \qquad (4)$$

Now since we have

$$r_t \in \mathbf{argmax}_{x \in U \setminus S_{t-1}=0} \frac{c^T x}{y^T Ax}$$

and since the items in $S'$, corresponding to the set $S^* \setminus S$, never are selected by the algorithm we must also have

$$(c^T r_t)(y_t^T A S') \geq (c^T S')(y_t^T A r_t) \qquad \forall t$$

Using this fact, (4) and (3), we get

$$\gamma \geq \frac{1}{V_{alg}} \sum_t (c^T S')(y_t^T A r_t) \geq \frac{1}{3V_{alg}} \sum_t (y_t^T A r_t)$$

We now use the fact that the updates to the dual variables are those of hedge algorithm (Kleinberg 2007). The hedge algorithm guarantees that the sum $\sum_t y^T A r_t$ will be almost as large as $\max_{y \ s.t \ \|y\|_1=1} y^T \sum_t A r_t$, up to minor errors terms. Using these guarantees gives

$$\gamma \geq \max_{y \ s.t \ \|y\|_1=1} \frac{1}{3V_{alg}} \left[ (1-\epsilon) \sum_t y^T A r_t - \frac{\log(m)}{\epsilon} \right]$$

Using the termination criteria of the algorithm, we get that $\max_{y \ s.t \ \|y\|_1=1} y^T \sum_t A r_t \geq \gamma - 1$ since no items are larger than 1 in any constraint (any such items cannot be part of any feasible solution by assumptions). Thus we have

$$\gamma \geq \frac{1}{3V_{alg}} \left[ \gamma - \epsilon\gamma - 1 - \frac{\log(m)}{\epsilon} \right]$$

Dividing both sides of the equation by $\gamma$ and using $\gamma = \beta \log(m)$ we get

$$1 \geq \frac{1}{3V_{alg}} \left[ 1 - \epsilon - \frac{1}{\log(m)\beta} - \frac{1}{\beta\epsilon} \right]$$

Now, choosing a sufficiently large $\beta$ and $\epsilon = \sqrt{1/\beta}$ (which optimizes the expression) completes the proof. Proof constants can be optimized for a slightly better bound, here avoided for clarity of presentation. ∎

Once a solution to the $\gamma$-width relaxed SUBSPPACK problem is obtained from Algorithm 1, we use the randomized Algorithm 2 to construct a feasible solution to our original problem, extending techniques from LP rounding to width-relaxed submodular programs (Bansal et al. 2012). The job of Algorithm 2 is to randomly remove items to achieve feasibility while guaranteeing that every item has a reasonable chance to not get removed. At a first stage, all but a fraction of all elements are thrown away independently and randomly, this guarantees that most constraints are likely to be satisfied. Secondly, every constraint is traversed in turn, and minor modifications based on items weights are done to ensure feasibility. The approximation ratio is proved below, a proof of feasibility given in the online version of the paper.

---

**Algorithm 2** $Rounding(\lambda, S)$
1: $S' \leftarrow \emptyset$
2: **for** item $s \in S$ **do**
3:     With probability $\lambda^{-1}$ let $S' \leftarrow S' \cup \{s\}$
4: **for** constraint $j = 1, 2...m$ **do**
5:     Let $S_j \subseteq S'$ denote set of elements $i'$ s.t. $A_{ji'} > 0$
6:     **for** Item $i \in S_j$ **do**
7:         **if** $\exists i' \neq i$ s.t. $i' \in S_j$ and $A_{ji'} > 1/2$ **then**
8:             Delete $i$ from $S'$
9:         **if** $\sum_{i' \in S_j s.t A_{ji'} < 1/2} A_{ji'} > 1$ **then**
10:            Delete $i$ from $S'$
11: **Return** $S'$

---

**Theorem 3** *Given a solution $S$ with value $V$ to the $\gamma$-width relaxation* SUBSPPACK, *with $\gamma = \mathcal{O}\big(\log(m)\big)$, Algorithm 2 with $\lambda = \mathcal{O}\big(\gamma k\big)$ produces a feasible solution to the original* SUBSPPACK *problem with expected value $\mathcal{O}\big(V \ /(k \ \gamma)\big)$.*

**Proof of apx. ratio.** The probability that any item gets removed is upper bounded by assuming that no other item has been removed previously. We let $E_{ij}$ denote the event that item $i$ gets removed from constraint $j$ in line 8 of Algorithm 2. We use $Y_j$ to denote the set $\{k \in S' | A_{jk} > 1/2\}$, by union bound and Markov's inequality we have

$$P[E_{ij} | i \in S] \leq \sum_{q \in Y_j} P[q \in S | i \in S] \leq \frac{|Y_j|}{\lambda}$$

Now, define $Y'_j$ as $\{k \in S' | A_{jk} \leq 1/2\}$ and denote the event that item $i$ is removed in line 10 of Algorithm 2 by $E'_{ij}$. Since $E'_{ij}$ implies $\sum_{q \in Y'_j \setminus i} A_{qj} > \frac{1}{2}$, we have $P[E'_{ij} | i \in S] \leq P[\sum_{q \in S_j \setminus i} A_{qj} > \frac{1}{2}]$. Again, we use the Markov inequality to bound this probability. Since every item in $Y_j$ has weight at least $1/2$, other items participating in constraint $j$ together take up space at most $\gamma - |Y_j|/2$. Hence, the Markov inequality we get

$$P[E'_{ij} | i \in S] \leq \frac{2}{\lambda}\left(\gamma - \frac{|Y_j|}{2}\right)$$

By union bound over the two ways items get removed, we see that the probability of item $i$ being discarded in constraint $j$ is less than $P[E_{ij}|i \in S] + P[E'_{ij}|i \in S] \leq \frac{2\gamma}{\lambda}$. Taking $\lambda = 4\gamma k$ and using union bounds over up $k$ constraints (recall that our program is $k$-columns sparse) gives us $P[i \text{ discarded}|i \in S] \leq \frac{1}{2}$. Now, imagine adding the items in the output of Algorithm 2 to a solution $S''$ in the order they were picked in Algorithm 1. The marginal utility $u_i$ of adding any item $i$ in $S''$ can by submodularity be lower bounded by its marginal utility $u'_i$ when it was picked in Algorithm 1. Since $\sum u'_i = V$ the approximation ratio now follows by linearity of expectations. ∎

**Proof of Theorem 1.** In order to combine Theorem 2 and 3 to get Theorem 1, we observe that any $\gamma$-relaxation of any SUBSPPACK problem can only increase OPT. The reason for this is that any solution satisfying $Ax \leq b$ also satisfies $Ax \leq \gamma b$, as we have $\gamma \geq 1$. Hence, Algorithm 1 gives us an constant factor solution $S$ to our original problem. By giving the solution $S$ with value $\mathcal{O}(\text{OPT})$ to Algorithm 2 we are guaranteed a feasible solution to our original problem with value $\mathcal{O}(\text{OPT}/k\gamma)$ by Theorem 3. Using $\gamma = \mathcal{O}(\log(m))$ completes the proof. ∎

### Performance Improvements

Our algorithm, like (Bansal et al. 2012), uses randomized rounding – which isn't the computational bottleneck in either method. By repeating it multiple times and taking the best result empirical performance can be improved, and as the optimal rounding parameter is problem specific, see (Shepherd and Vetta 2007), a parameter-sweep can be used. To avoid unnecessary evaluations of $f(X)$ greedy algorithms can use the standard technique of 'lazy evaluations' (Minoux 1978), while algorithms like (Bansal et al. 2012) can be sped up by simply "memorizing" evaluations.

## Experiments

### Biocontrol of the Hemlock Wolly Adelgid

As part of an ongoing collaboration with the Department of Natural Resources at Cornell, we evaluate our method in the context of biocontrol of the invasive species Hemlock Wolly Adelgid (HWA), see Figure 4. First observed in the 1950s near Richmond, Virginia, the insect pest has spread to nineteen eastern states causing wide-ranging destruction to native hemlock trees (Preisser, Oten, and Hain 2014). The scenario we model is citizen science-based biocontrol, where volunteers release the HWA predator *Sasajiscymnus tsugae*. Citizens scientists, volunteers in scientific work, present an opportunity for substantial cost savings for conservation organizations (Conrad and Hilchey 2011), and conservation land trusts have already experimented with using private citizens for biocontrol (Horan 2006). A major barrier to using citizen scientists is the distance they are willing to travel (Xue et al. 2016). Many ideal locations to release the predator is far from residential areas and the total logistic effort of citizen scientists constrains conservation strategies. The insect predator *Sasajiscymnus tsugae* is known to only attack HWA in nature (Cheah and McClure ) as per the assumption in our model, but has to be artificially bred (Onken and Richard 2011). The process is expensive, and the number of available predators presents the second constraint.

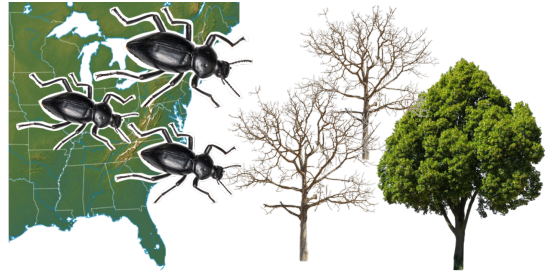

Figure 4: The HWA is an invasive insect pest that suck nutrients from twigs of the hemlock trees, eventually killing the host tree (Orwig and Foster 1998). The Hemlock is a foundation species in the eastern North America providing critical shelter and habitat upon which a whole range of species depend (Quimby and others 1996), and biocontrol of HWA is deemed crucial to the continued viability of hemlocks (Onken and Richard 2011).

### Modelling the Biocontrol Problem

The eastern US is modeled as a network of nodes corresponding to 20-by-20 kilometer patches in a grid layout. Hemlock occurs in $3489$ nodes which participate in the cascades. Every patch has an associated elevation, distribution of winter temperatures, mean temperature and hemlock abundance from ecological and geographical datasets (Fitzpatrick et al. 2012) (Rohde et al. 2013). The model for the HWA is based on the spatially explicit metapopulation model for HWA introduced by (Fitzpatrick et al. 2012). Following this work we take edge transmission probability $P(i \text{ spreads to } j)$ equal to $P(i \text{ disperses to } j)P(\text{insect establishes in } j)$ for both predator and prey insects, using its log-normal distribution for the dispersal probability. Food scarcity and winter mortality are both important for HWA population dynamics (Paradis et al. 2008), drawing upon (Fitzpatrick et al. 2012) $P(\text{HWA establishes in } j)$ is proportional to the number of hemlock trees in location $j$ and $0.507 - 0.078T_j$ where $T_j$ is the average winter temperature(°C) sampled at $j$. We take the predator establishment probability proportional to the logistic regression model from the empirical work of (Hakeem et al. 2013), depending on elevation and temperature. For the HWA the constant of proportionality is taken to be $680.0$ to roughly match historical data (Morin, Liebhold, and Gottschalk 2009) (Fitzpatrick et al. 2012). The total number of edges in the graph, using the above edge weights and removing edges with vanishingly small weight, is 110525 and 442457 for the predator and prey respectively. We model two resource constraints: the number of available predator insects and the total effort of participating citizen scientists. For any year the number of required insects to establish a population at location $j$ is assumed to be inversely proportional to $P(\text{insect establishes in } j)$. Exact costs for citizen scientists are hard to estimate; as a proxy, we use the distance from the the target location to any major city (popu-lation $\geq 100{,}000$), plus 30km representing baseline effort. Experiments with different budget parameters and predator constant of proportionality are given in the online version of the paper (Bjorck et al. 2017). The results presented in the main text take the predator constant of proportionality as 2 (roughly matching the HWA virality), the ratio between the citizen science and predator budgets as 37.5 (both constraints being roughly equally important) and a total budget that allows the selection of at most $0.1\%$ of the network nodes. The simulation runs 1951-2051 with initial conditions of (Fitzpatrick et al. 2012) and biocontrol efforts 2021-2027. We use two-year time steps, roughly corresponds to the timescale from initial to complete infestation in (Fitzpatrick et al. 2012).

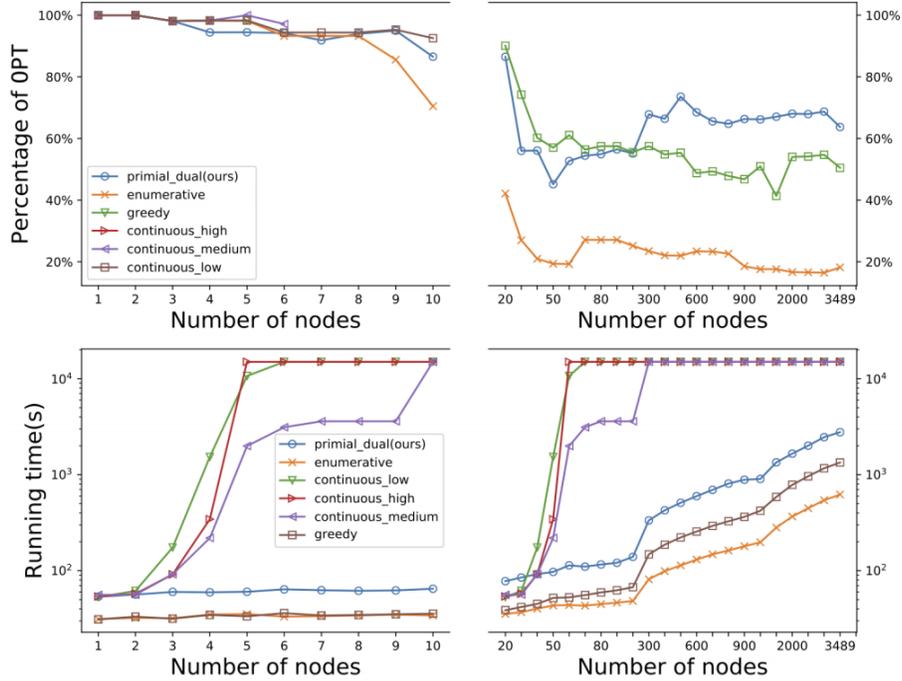

Figure 5: Algorithm performance is studied as the problem size is varied. Note that the parameter range is nonlinear. (Upper) Our algorithm provides the best alternative for large-scale instances as the high performing continuous algorithm times out. (Lower) The continuous algorithms times out for small instances, while other algorithms scales within a constant of each other. `OPT` is upper bounded, and algorithms are potentially better than shown.

### Experimental Setup

We compare our method against the $\mathcal{O}(k)$ approximation algorithm of (Bansal et al. 2012). It relies on randomized rounding and the **continuous** relaxation $f(s) = \sum_{R \subseteq U} f(R) \prod_{i \in R} s_i \prod_{j \notin R}(1 - s_i)$ which agrees with the original objective function on integral points. The algorithm essentially performs gradient descent along the solution to a LP with objective $\nabla f(s)$, where $\nabla f(s)$ is obtained by sampling. We consider three sampling rates: the original high rate $n^5$, medium $n^3$ and low $n$. A second baseline is the fast algorithm of (Badanidiyuru and Vondrák 2014) which chooses items with marginal utility over combined size in all constraints above a threshold that is **enumerated** with multiplicative spacing $\epsilon$ (we take $\epsilon = 0.1$). It runs in $\mathcal{O}(mn \log^2 n)$ but only has an approximation ratio of $\mathcal{O}(m)$. These two algorithms represent the state of the art in provable methods. As a last baseline, we implement the greedy algorithm that provides no guarantees but can nonetheless work well. It iteratively picks the item with the highest marginal utility that can feasibly be added, until no more items can be added. Our method uses $\beta = 7$, based on empirical performance, although it does provide a guarantee through Theorem 1. The algorithm is, however, stable with regards to this parameter, see the online version of the paper for the relevant parameter sensitivity experiments.

As is practice in stochastic optimization we approximate the expectation (1) using the average of a number (250) of samples drawn independently. The methodology is known as sample average approximation and comes with asymptotic optimality guarantees (Shapiro 2003). All algorithms are implemented in c++ and the evaluation of (1) parallelized. The LPs are solved with the commercial software CPLEX. Repeated randomization (500 times), lazy evaluations, and memoization are used according to earlier sections as applicable. Experiments are run on cluster nodes outfitted with 12 cores on two Intel x5690 processors and 48GB RAM.

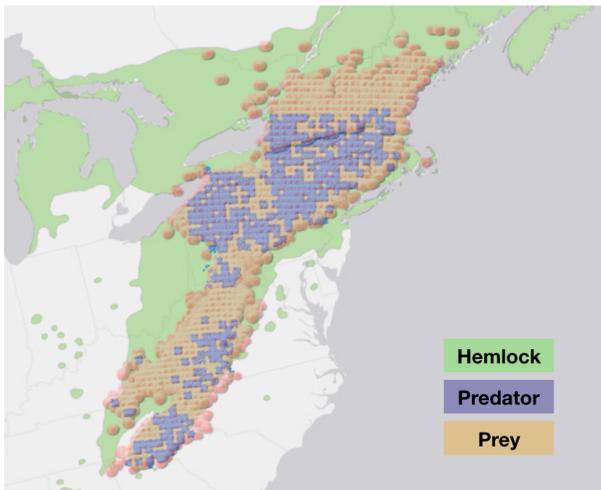

Figure 6: The map illustrates an outcome of the strategy suggested by our algorithm at year 2051. The HWA has continued to spread north, this region, rich in hemlock, is the primary focus of conservation efforts. The middle region of the study area gets little protection, its hemlock population is relatively sparse and a predator introduced there can only spread in two directions.

## Experimental Results

We first study the scalability and performance of the algorithms by restricting them to an increasingly large randomly selected subset of the nodes of the complete network. The mean solution quality and running time (wall clock) of ten runs, using 4 hours timeout, are reported in Figure 5. `OPT` is upper bounded by techniques of (Leskovec et al. 2007). Our algorithm outperforms competitors, the continuous algorithm scales poorly and times out even for medium-sized problems while the fast enumerative algorithm never achieves good performance. Small instances are relatively easy problems as there are only a handful 'good' nodes to chose from, the greedy algorithm slightly outperforms our method for these but performs significantly worse for larger instances. These results are robust as problem parameters are varied, see the online version (Bjorck et al. 2017) for parameter experiments. Figure 6 shows the effects of the management strategy chosen by our method. Our algorithm provides a relatively non-uniform strategy, focusing primarily on northern parts. The middle part of the eastern North America, where the hemlock forest is relatively sparse there, is largely ignored, pointing towards non-trivial management strategies.

## Related Work

Invasive species management and predator-prey dynamics are studied broadly, from stylized dynamic systems models for spatially homogeneous systems (Sun, Zhang, and Tian 2016) to reinforcement learning for small-scale problems (Taleghan et al. 2015). Computational work for large-scale spatially explicit models commonly employs stochastic dynamic programming which cannot express combinatorial management strategies (Shea and Possingham 2000) or MIP programming with poor scalability and no guarantees (Büyüktahtakın, Feng, and Szidarovszky 2014). MIP programming has been successful broadly in computational ecology (Dilkina and Gomes 2010) (Sheldon et al. 2012), the authors experimented with MIP formulations for our biocontrol problem but found that it didn't scale well. The more algorithmic perspective we employ instead provide guarantees, useful future biocontrol applications with currently unknown parameters. The only work on biocontrol within computer science known to the authors is the purely theoretical (Spencer 2012), which is primarily focused on reductions and extensions of the firefighter problem (Elliot Anshelevichm Deeparnab Chakrabarty 2009).

## Conclusions

We have formulated a novel problem of optimal biocontrol and shown that it is expressible as submodular optimization problem subject to sparse packing constraints. We have developed a novel relaxation-based algorithm for this broad problem class that both theoretically and empirically outperforms, or out scales, competing methods. This has allowed us to find near-optimal strategies for biocontrol of the HWA, a problem of great ecological and economic importance.

**Acknowledgements.** The authors would like to thank the anonymous reviewers, Matthew Fitzpatrick, Saskya van Nouhuys, Rich Bernstein, Jennifer Dean, Carrie Jean Brown-Lima and Angela Fuller. This research was supported by the National Science Foundation (CCF-1522054 and CNS-1059284), New York State Department of Environmental Conservation (Award C008698 among others) and USDA Forest Service.

## References

Aspnes, J.; Chang, K.; and Yampolskiy, A. 2005. Inoculation strategies for victims of viruses and the sum-of-squares partition problem. In *SODA*.

Badanidiyuru, A., and Vondrák, J. 2014. Fast algorithms for maximizing submodular functions. In *SODA*.

Badanidiyuru, A.; Mirzasoleiman, B.; Karbasi, A.; and Krause, A. 2014. Streaming submodular maximization: Massive data summarization on the fly. In *KDD*.

Bansal, N.; Korula, N.; Nagarajan, V.; and Srinivasan, A. 2012. Solving packing integer programs via randomized rounding with alterations. *Theory of Computing*.

Bjorck, J.; Bai, Y.; Wu, X.; Yexiang, X.; Whitmore, M.; and Gomes, C. 2017. Scalable relaxations of sparse packing constraints: Optimal biocontrol in predator-prey networks. *arXiv*.

Burnett, S. 1997. Colonizing cane toads cause population declines in native predators: reliable anecdotal information and management implications. *Pacific Conservation Biology*.

Büyüktahtakın, İ. E.; Feng, Z.; and Szidarovszky, F. 2014. A multi-objective optimization approach for invasive species control. *Journal of the Operational Research Society*.

Calinescu, G.; Chekuri, C.; Pál, M.; and Vondrák, J. 2007. Maximizing a submodular set function subject to a matroid constraint. In *IPCO*.

Cheah, C., and McClure, M. S. Sasajiscymnus (formerly pseudoscymnus) tsugae. https://biocontrol.entomology.cornell.edu/predators/sasajiscymnus.php. Accessed: 2017-07-30.

Conrad, C. C., and Hilchey, K. G. 2011. A review of citizen science and community-based environmental monitoring: issues and opportunities. *Environmental monitoring and assessment*.

Cormen, T. H. 2009. *Introduction to algorithms*.

Dilkina, B. N., and Gomes, C. P. 2010. Solving connected subgraph problems in wildlife conservation. In *CPAIOR*, volume 6140, 102–116. Springer.

Elliot Anshelevichm Deeparnab Chakrabarty, Ameya Hate, C. S. 2009. Approxmation algorithms for the firefighter problem: Cuts over time and submodularity. In *International Symposium on Algorithms and Computation*.

Fitzpatrick, M. C.; Preisser, E. L.; Porter, A.; Elkinton, J.; and Ellison, A. M. 2012. Modeling range dynamics in heterogeneous landscapes: invasion of the hemlock woolly adelgid in eastern north america. *Ecological Applications*.

Garey, M. R., and Johnson, D. S. 2002. *Computers and intractability*, volume 29. wh freeman New York.

Goldschmidt, T.; Witte, F.; and Wanink, J. 1993. Cascading effects of the introduced nile perch on the detritivorous/phytoplanktivorous species in the sublittoral areas of lake victoria. *Conservation Biology*.

Gomes, C. P. 2009. Computational sustainability: Computational methods for a sustainable environment, economy, and soc iety. *The Bridge*.

Hakeem, A.; Grant, J.; Wiggins, G.; Lambdin, P.; Hale, F.; Buckley, D.; Rhea, J.; Parkman, J.; and Taylor, G. 2013. Factors affecting establishment and recovery of sasajiscymnus tsugae. *Environmental entomology*.

Hanski, I. 1999. *Metapopulation ecology*.

Hastings, A. 1977. Spatial heterogeneity and the stability of predator-prey systems. *Theoretical population biology*.

Hazan, E.; Safra, S.; and Schwartz, O. 2006. On the complexity of approximating k-set packing. *computational complexity*.

Hoi, S. C.; Jin, R.; Zhu, J.; and Lyu, M. R. 2006. Batch mode active learning and its application to medical image classification. In *ICML*.

Horan, P. 2006. Community-based hwa predator beetle releases. http://savinghemlocks.org/community_diy_predator-beetle-releases_sasi_st/. Accessed: 2017-07-30.

Kallrath, J., and Schreieck, A. 1995. Discrete optimisation and real world problems. In *High-performance computing and networking*.

Kempe, D.; Kleinberg, J.; and Tardos, É. 2003. Maximizing the spread of influence through a social network. In *KDD*.

Kleinberg, R. 2007. Lecture notes for cs 683, cornell university, 2007. http://www.cs.cornell.edu/courses/cs683/2007sp/lecnotes/week2.pdf.

Leskovec, J.; Krause, A.; Guestrin, C.; Faloutsos, C.; VanBriesen, J.; and Glance, N. 2007. Cost-effective outbreak detection in networks. In *KDD*.

Minoux, M. 1978. Accelerated greedy algorithms for maximizing submodular set functions. *Optimization Techniques*.

Morin, R. S.; Liebhold, A. M.; and Gottschalk, K. W. 2009. Anisotropic spread of hemlock woolly adelgid in the eastern united states. *Biological Invasions*.

Onken, B., and Richard, R. 2011. *ImplementatIon and status of biological control of the hemlock woolly adelgid*. US Forest Service.

Orwig, D. A., and Foster, D. R. 1998. Forest response to the introduced hemlock woolly adelgid in southern new england, usa. *Journal of the Torrey Botanical Society*.

Paradis, A.; Elkinton, J.; Hayhoe, K.; and Buonaccorsi, J. 2008. Role of winter temperature and climate change on the survival and future range expansion of the hemlock woolly adelgid (adelges tsugae) in eastern north america. *Mitigation and Adaptation Strategies for Global Change*.

Pimentel, D.; Zuniga, R.; and Morrison, D. 2005. Update on the environmental and economic costs associated with alien-invasive species in the united states. *Ecological economics*.

Plotkin, S. A.; Shmoys, D. B.; and Tardos, É. 1995. Fast approximation algorithms for fractional packing and covering problems. *Mathematics of Operations Research*.

Preisser, E. L.; Oten, K. L.; and Hain, F. P. 2014. Hemlock woolly adelgid in the eastern united states: What have we learned?

Quimby, J., et al. 1996. Value and importance of hemlock ecosystems in the eastern united states. *Proceedings of the first hemlock woolly adelgid review*.

Rendell, P. 2002. Turing universality of the game of life. In *Collision-based computing*.

Rohde, R.; Muller, R.; Jacobsen, R.; Perlmutter, S.; Rosenfeld, A.; Wurtele, J.; Curry, J.; Wickhams, C.; and Mosher, S. 2013. Berkeley earth temperature averaging process, geoinfor. geostat.: An overview 1 : 2. *of* 13:20–100.

Sasaji, H., and McClure, M. S. 1997. Description and distribution of pseudoscymnus tsugae. *Annals of the Entomological Society of America*.

Shapiro, A. 2003. Monte carlo sampling methods. *Handbooks in operations research and management science*.

Shea, K., and Possingham, H. P. 2000. Optimal release strategies for biological control agents: an application of stochastic dynamic programming to population management. *Journal of Applied Ecology*.

Sheldon, D.; Dilkina, B.; Elmachtoub, A. N.; Finseth, R.; Sabharwal, A.; Conrad, J.; Gomes, C. P.; Shmoys, D.; Allen, W.; Amundsen, O.; et al. 2012. Maximizing the spread of cascades using network design. *arXiv*.

Shepherd, F. B., and Vetta, A. 2007. The demand-matching problem. *Mathematics of Operations Research*.

Spencer, G. 2012. Robust cuts over time: Combatting the spread of invasive species with unreliable biological control. In *AAAI*.

Sun, K.; Zhang, T.; and Tian, Y. 2016. Theoretical study and control optimization of an integrated pest management predator–prey model with power growth rate. *Mathematical biosciences*.

Sviridenko, M. 2004. A note on maximizing a submodular set function subject to a knapsack constraint. *Operations Research Letters*.

Taleghan, M. A.; Dietterich, T. G.; Crowley, M.; Hall, K.; and Albers, H. J. 2015. Pac optimal mdp planning with application to invasive species management.

Xue, Y.; Davies, I.; Fink, D.; Wood, C.; and Gomes, C. P. 2016. Behavior identification in two-stage games for incentivizing citizen science exploration. In *CP*.

# Appendix

**Proof Sketch for proposition 1.** A simple reduction comes from the maximum set cover problem (Garey and Johnson 2002) through constructing a graph where spreading the predator becomes equivalent to covering all elements. Given a maximum set cover problem over universe $U$ with sets $\{S_i\}_{i=1}^{T}$ and budget $k$ we create a graph $G$. For every element $u \in U$ we create a node, and for every set $S_t$ we create a node and connect it with elements covered by the corresponding set with edge probability 1. All nodes are infected at $t=0$ and the budgets is set such we can only chose $k$ out of the $T$ "set-nodes", solving this problem then entails solving the corresponding maximum set cover problem. ∎

**Proof of feasiblity in Theorem 3** Consider the set $S'$ of items when the algorithm terminates and any single constraint $m$. If any item is 'big' in $m$ we cannot have any other items with nonzero coefficient in $m$ by the design of the algorithm, and a single 'big' item is feasable in constraint $m$. If we do not have items 'big' in $m$ the sum of the weight of all small items is at most 1 by the design of the algorithm, this again is feasable in $m$ and since this holds for any constraint the whole solution is feasable. ∎

**Proof sketch for proposition 2.** As in (Kempe, Kleinberg, and Tardos 2003) we can imagine the random events corresponding to whether an edge conducts spread or not at some timestep $t$ as being specified in advance. Let us construct an unweighted time-layered graph $G' = (V', E')$ as (Kempe, Kleinberg, and Tardos 2003) for the predator. The node set is $V' = V \times T$, any two nodes $(v,t)$ and $(v,t+1)$ are connected, and for any edge $(u,v)$ that transmitts at time $t$ we have an edge between nodes $(v,t)$ and $(u,t+1)$. The prey can spread from nodes not unoccupied (recall the predator state implies a balanced mixed population), for the prey we can thus construct a similar graph $G'' = (V'', E'')$ representing spread from either prey or predator state. Through this graph we can define $I(t) \subseteq V \times T$ as the set of nodes in the prey state at time $t$ of the graph given any fixed starting set $I(0)$ using distance $d$ (the number of edges between nodes) in the graph as $I(t) = \{v \in V | \exists v' \in I(0) \ s.t. \ d_{G''}(v,v') \le t\} \times \{t\}$. The expectation over all different random events then becomes the linear combination over all graphs $G', G''$, and possibly random starting conditions weighted by their probabilities of the type $\prod_{e \in E'} p_e \prod_{e \notin E'}(1 - p_e)$. It now suffices to show that the function over one combination of these graphs is monotonically submodular. We let $I$ be the union of $I(0), I(1)$ and so on, and note that node $i$ becomes saved at time $t'$ by introducing the predator at node $j$ at time $t$ if and only if $\exists t'' < t' \ s.t. \ i \in I(t'') \wedge \exists$ path $p(j \to i) \subseteq E' \cup I \ s.t. \ length(p) = t'' - t$. Path here is a set of nodes and edges connecting them, and its length the number of edges. We see that every node has a hitting set, and hitting set functions are monotonically submodular. ∎

## Robustness experiments

We choose multiple parameters of percentage of total budges and the predator virality to test the robustness of our algorithm. Here we consider the whole network(i.e the algorithms need to consider the whole nodes set), where the **continuous** algorithm cannot scale. In Figure 7, we could see that our algorithm performs the best at all the time, which leads to excellent robustness.

## Constraint-wise experiments

We conduct two experiments the individual budgets of our two constraints. Figure 8 shows how different budgets impacts the performance of different algorithms, and highlights the robustness of our method.

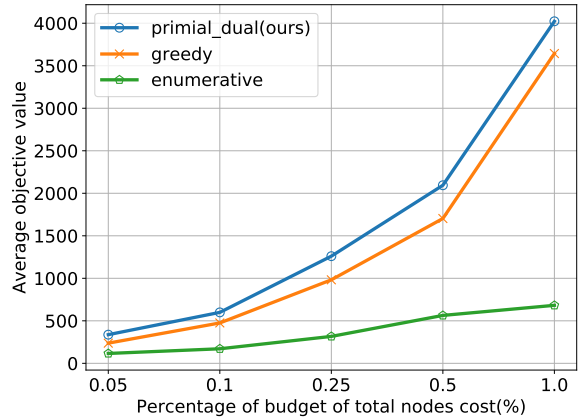

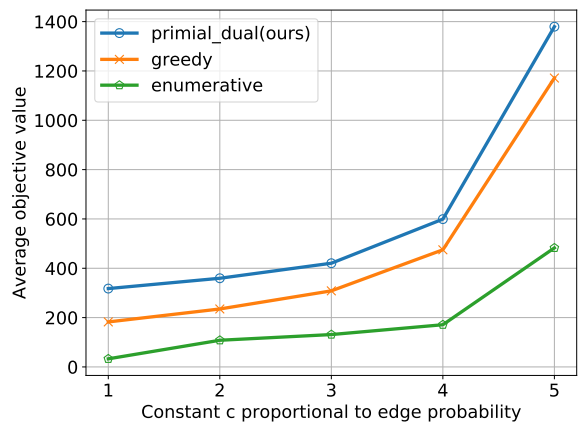

Figure 7: We here illustrate how algorithm performance depends on budgets (the % of coverable nodes) and predator virality (constant of proportionality for edge transmission). Our method outperforms other methods and shows robustness to parameters.

## Parameter sensitivity

We tested the sensitivity of our algorithm to the choice of $\beta$. In general the optimal value should be problem specific, for our problem the algorithm performed well for a broad range of values, as illustrated in Figure 9.

## Datasets

We obtained the dataset for the distribution of hemlock trees from M. Fitzpatrick and will distribute the raw data with his permission. The elevation data was obtained through the Google maps API, the temperature data from Berkeley earth (Rohde et al. 2013) and (Fitzpatrick et al. 2012).

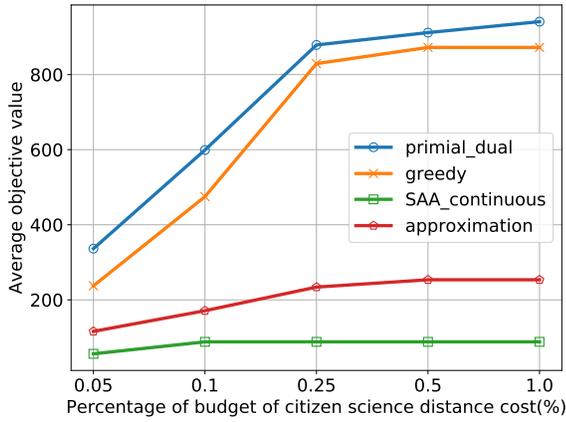

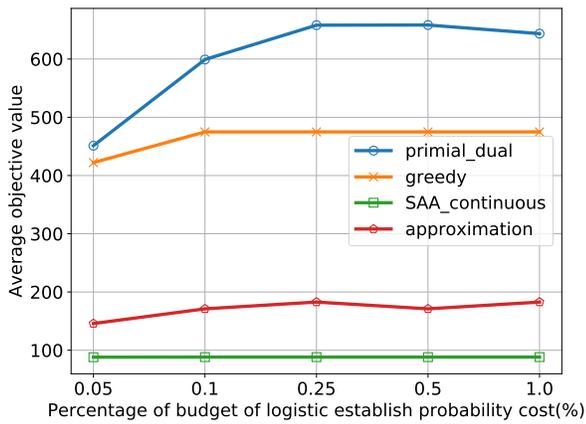

Figure 8: We here illustrate how algorithm performance depends on citizen science budgets(Upper) and predator budgets(Lower). We keep the fixed constraint as 0.1% budget of total nodes cost when we modify another constraints. Our method outperforms other methods and shows robustness to parameters.

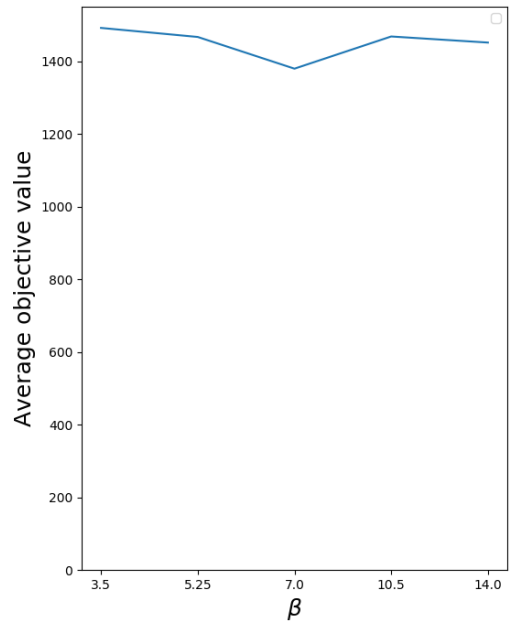

Figure 9: We here illustrate the sensitivity of our algorithms performance with respect to the parameter $\beta$, for the complete network. The effect of changing $\beta$ is relatively small across the whole range, which suggests that the algorithm is roboust and doesn't require perfect tuning.